\documentclass{ieeeaccess}
\usepackage{cite}
\usepackage{amsmath,amssymb,amsfonts}
\usepackage{graphicx}
\usepackage{textcomp}
\def\BibTeX{{\rm B\kern-.05em{\sc i\kern-.025em b}\kern-.08em
    T\kern-.1667em\lower.7ex\hbox{E}\kern-.125emX}}

\usepackage{subcaption}
\usepackage{hyperref}
\usepackage{url}
\usepackage{booktabs}       % professional-quality tables
\usepackage{nicefrac}       % compact symbols for 1/2, etc.
\usepackage{microtype}      % microtypography
\usepackage{color, colortbl}
\usepackage{pifont}% http://ctan.org/pkg/pifont
\newcommand{\cmark}{\ding{51}}%
\newcommand{\xmark}{\ding{55}}%
\usepackage{multirow,}
\usepackage{algorithm,algorithmicx,algpseudocode}

\begin{document}
\history{Date of publication December 24, 2019, date of current version January 03, 2020.}
\doi{10.1109/ACCESS.2019.2961960}

\title{A Low Effort Approach to Structured CNN Design Using PCA}
\author{\uppercase{Isha Garg},
\uppercase{Priyadarshini Panda, and Kaushik Roy
}}
\address{School of Electrical and Computer Engineering, Purdue University, West Lafayette, IN 47907, USA}
\tfootnote{This work was supported in part by the Center for Brain Inspired Computing (C-BRIC), one of the six centers in JUMP, a Semiconductor Research Corporation (SRC) program sponsored by DARPA, by the Semiconductor Research Corporation, the National Science Foundation, Intel Corporation, the DoD Vannevar Bush Fellowship, and by the U.S. Army Research Laboratory and the U.K. Ministry of Defence under Agreement Number W911NF-16-3-0001. }

\markboth
{I. Garg \headeretal: A Low Effort Approach to Structured CNN Design Using PCA}
{I. Garg \headeretal: A Low Effort Approach to Structured CNN Design Using PCA}

\corresp{Corresponding author: Isha Garg (e-mail: gargi@purdue.edu).}

\begin{abstract}
Deep learning models hold state of the art performance in many fields, yet their design is still based on heuristics or grid search methods that often result in overparametrized networks. This work proposes a method to analyze a trained network and deduce an optimized, compressed architecture that preserves accuracy while keeping computational costs tractable. Model compression is an active field of research that targets the problem of realizing deep learning models in hardware. However, most pruning methodologies tend to be experimental, requiring large compute and time intensive iterations of retraining the entire network. We introduce structure into model design by proposing a single shot analysis of a trained network that serves as a first order, low effort approach to dimensionality reduction, by using PCA (Principal Component Analysis). The proposed method simultaneously analyzes the activations of each layer and considers the dimensionality of the space described by the filters generating these activations. It optimizes the architecture in terms of number of layers, and number of filters per layer without any iterative retraining procedures, making it a viable, low effort technique to design efficient networks. We demonstrate the proposed methodology on AlexNet and VGG style networks on the CIFAR-10, CIFAR-100 and ImageNet datasets, and successfully achieve an optimized architecture with a reduction of up to 3.8X and 9X in the number of operations and parameters respectively, while trading off less than 1\% accuracy. We also apply the method to MobileNet, and achieve 1.7X and 3.9X reduction in the number of operations and parameters respectively, while improving accuracy by almost one percentage point.
\end{abstract}

\begin{keywords}
CNNs, Efficient Deep Learning, Model Architecture, Model Compression, PCA, Dimensionality Reduction, Pruning, Network Design
\end{keywords}

\titlepgskip=-15pt

\maketitle

\section{Introduction}
\IEEEPARstart{D}{eep} Learning is widely used in a variety of applications, but  often suffers from issues arising from exploding computational complexity due to the large number of parameters and operations involved. With the increasing availability of compute power, state of the art Convolutional Neural Networks (CNNs) are growing rapidly in size, making them prohibitive to deploy in energy-constrained environments. This is exacerbated by the lack of a principled, explainable way to reason out the architecture of a neural network, in terms of the number of layers and the number of filters per layer. In this paper, we refer to these parameters as the depth and layer-wise width of the network, respectively. The design of a CNN is currently based on heuristics or grid searches for optimal parameters \cite{benjio-prac}. Often, when a designer wants to develop a CNN for new data, transfer learning is used to adapt well-known networks that hold state of the art performance on established datasets. This adaptation comes in the form of minor changes to the final layer and fine-tuning on the new data. It is rare to evaluate the fitness of the original network on the given dataset on fronts other than accuracy. Even for networks  designed from scratch, it is common to either perform a grid search for the network architecture, or to start with a variant of 8-64 filters per layer, and double the number of filters per layer as a rule of thumb \cite{VGG}, \cite{resnet}. This often results in an over-designed network, full of redundancy \cite{Denil}. Many works have shown that networks can be reduced to a fraction of their original size without any loss in accuracy \cite{han2015deep}, \cite{lecun1990optimal}, \cite{denton2014exploiting}. This redundancy not only increases training time and computational complexity, but also creates the need for specialized training in the form of dropout and regularization \cite{Dropout}.

% \begin{figure*}[!t]
% \centering
% \subfloat[Case I]{\includegraphics[width=2.5in]{box}%
% \label{fig_first_case}}
% \hfil
% \subfloat[Case II]{\includegraphics[width=2.5in]{box}%
% \label{fig_second_case}}
% \caption{Simulation results for the network.}
% \label{fig_sim}
% \end{figure*}

% \begin{figure*}[!t]
% \centering
% \subfloat{\includegraphics[width=1\textwidth]{Images/Standard_Pruning.png}%
% \caption{}
% \label{fig:std_cht}}
% \hfil
% \subfloat{\includegraphics[width=1\textwidth]{Images/PCA_Pruning.png}%

% \caption{}
% \label{fig:pca_cht}}
% \caption{Fig. (\ref{fig:std_cht}) shows the flowchart for standard pruning techniques. It involves two multiplicative loops, each involving retraining of the entire network. In contrast, our technique, shown in Fig. (\ref{fig:pca_cht}) only requires a single retraining iteration for the whole network.}
% \label{fig:charts}
% \end{figure*}

\begin{figure*}[!t]
\centering
\begin{subfigure}{1\textwidth}
\centering
\includegraphics[width=1\textwidth]{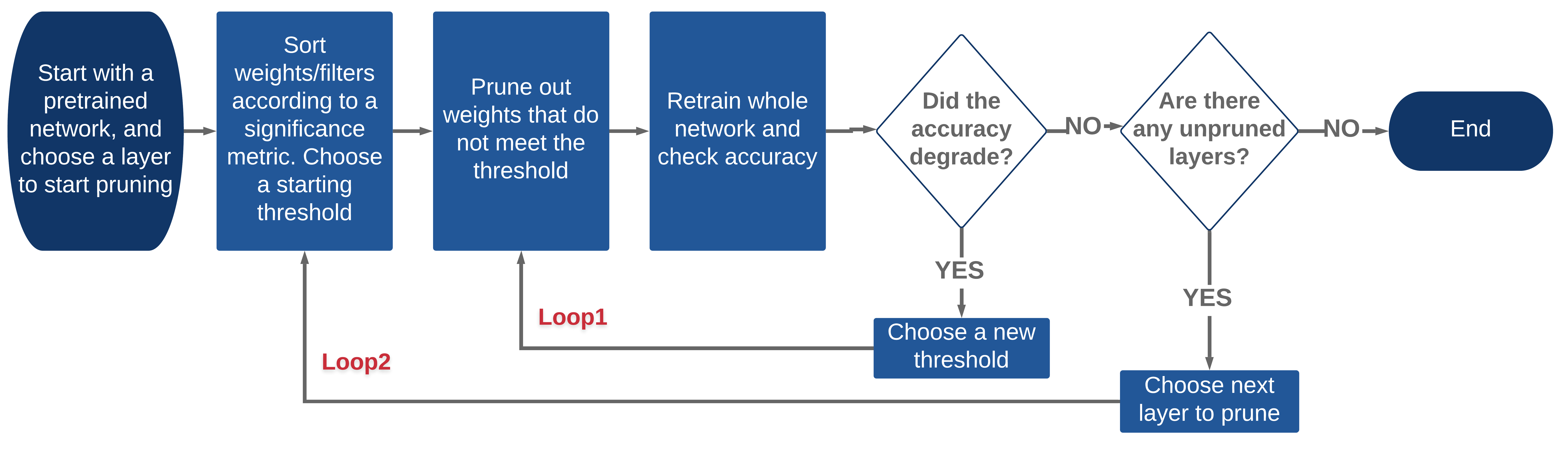} 
\caption{}
\label{fig:std_cht}
\end{subfigure}
\centering
\begin{subfigure}{1\textwidth}
\centering
\includegraphics[width=1\textwidth]{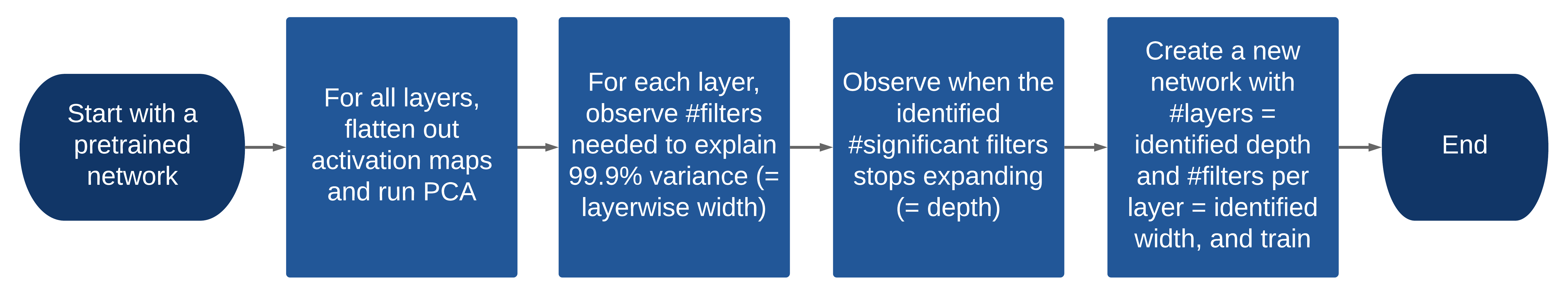}  
\caption{}
\label{fig:pca_cht}
\end{subfigure}
\caption{Fig. \ref{fig:std_cht} shows the flowchart for standard pruning techniques. It involves two multiplicative loops, each involving retraining of the entire network. In contrast, the proposed technique, shown in Fig. \ref{fig:pca_cht} only requires a single retraining iteration for the whole network.}
\label{fig:charts}
\end{figure*}

\par \textbf{Practical Problems with Current Model Compression Methods:} The field of model compression explores ways to prune a network post training in order to remove redundancy. However, most of these techniques involve multiple time and compute intensive iterations to find an optimal threshold for compression, making it impractical to compress large networks \cite{lecun1990optimal}, \cite{comp-aware}, \cite{jaderberg2014speeding}. An iteration here is referred to as the entire procedure of training or retraining a network, instead of a forward and backward pass on a mini-batch. Most standard pruning techniques broadly follow the methodology outlined in the flowchart in Fig. \ref{fig:std_cht}. They start with a pre-trained network and prune the network layer by layer, empirically finding a threshold for pruning in each layer. The pruning threshold modulates the fraction of pruning performed at each iteration and that, in turn, affects the accuracy, which is estimated by retraining. This results in the two loops shown in the figure. Loop 1 iterates to find a suitable pruning threshold for a layer, and Loop 2 repeats the entire process for each layer. Since these loops are multiplicative, and each iteration involves retraining the whole network, pruning a network becomes many times more time and compute intensive than training it. Some methods require only one of the two loops \cite{molchanov2016pruning}, \cite{nisp}, but that still results in a large number of retraining iterations for state of the art networks. Furthermore, the resulting thresholds are not explainable, and therefore can not usually be justified or predicted with any degree of accuracy. 

\par \textbf{Proposed Method to Optimize Architecture:} To address these issues, we propose a low effort technique that uses Principal Component Analysis (PCA) to analyze the network in a single  pass, and gives us an optimized design in terms of the number of filters per layer (width), and the number of layers (depth) without the need for retraining. Here, we refer to optimality in terms of removal of redundancy. We do not claim that our method results in the most optimal architecture. However, it is, to the best of our knowledge, a method which optimizes a pre-trained network with the lowest effort in terms of retraining iterations. The proposed method is elucidated in Fig. \ref{fig:pca_cht}. We start with a pre-trained network, and analyze the activations of all layers simultaneously using PCA. We then determine the optimized network's layer-wise width from the number of principal components required to explain 99.9\% of the cumulative explained variance. We call these the `significant dimensions' of each layer and optimize the depth based on when these significant dimensions start contracting. Once the requisite width and depth are identified, the user can create a new, randomly initialized network of the identified width and depth and train once to get the final, efficient model. It removes both the loops since we analyze the entire network in one shot, and have a pre-defined threshold for each layer instead of an empirical one. The proposed method optimizes the architecture in one pass, and only requires a total of one retraining iteration for the whole network, drastically reducing the time required for compression. In addition, the choice of the threshold is predetermined and explainable, and therefore can be adapted to suit different energy budgets. This provides an accuracy-efficiency tradeoff knob that can be utilized for more error-tolerant applications where energy consumption is a driving factor in model architecture search. 

\par \textbf{Contributions:} The main contribution of this work is a practical compression technique with explainable design heuristics for optimizing network architecture, at negligible extra compute cost or time. To the best of our knowledge, this is the first work that analyzes all layers of networks simultaneously and optimizes structure in terms of both width and depth, without any iterative searches for thresholds for compression per layer. The additional benefits of using this methodology are two-fold. First, for more error tolerant applications, where accuracy can sometimes be traded for faster inference or lower energy consumption, this analysis offers a way to gracefully tune that trade-off. Second, the resultant PCA graphs (Fig. \ref{fig:sensitivity}) are indicative of the sensitivity of layers and help identify layers that can be aggressively targeted while compressing the network. This is discussed in detail in section III. The effectiveness of the proposed methodology to optimize the structures of some widely used network architectures is demonstrated in Section IV.

\begin{table}[]
\begin{tabular}{|>{\centering\arraybackslash}m{2.65cm}|>{\centering\arraybackslash}m{.8cm}|>{\centering\arraybackslash}m{.8cm}|>{\centering\arraybackslash}m{1.1cm}|>{\centering\arraybackslash}m{1.35cm}|}
\hline
Method Name                                          & Loop1 & Loop2  &  Custom Training & Custom Architecture/ Hardware  \\ \hline
Deep Compression\cite{han2015deep} & \cmark                    & \cmark       & \xmark    & \cmark                \\ \hline
OBD \cite{lecun1990optimal}                                 & \cmark                    & \xmark       & \xmark    & \cmark                \\ \hline
OBS \cite{optimalbrainsurgeon}                               & \cmark                    & \xmark       & \xmark    & \cmark                \\ \hline
Denton \cite{denton2014exploiting}                                              & \cmark                    & \cmark       & \xmark    & \cmark                \\ \hline
Comp-Aware\cite{comp-aware}                                   & \xmark                    & \cmark       & \cmark    & \cmark                \\ \hline
Jaderberg \cite{jaderberg2014speeding}                                           & \cmark                    & \xmark       & \cmark    & \xmark                \\ \hline
Molchanov \cite{molchanov2016pruning}                                           & \cmark                    & \cmark       & \xmark    & \xmark                \\ \hline
NISP \cite{nisp}                                                & \cmark                    & \xmark       & \xmark    & \xmark                \\ \hline
AxNN    \cite{AxNN}                                             & \cmark                    & \cmark       & \xmark    & \cmark                \\ \hline
Li\_efficient \cite{li2016pruning}                              & \cmark                    & \cmark       & \xmark    & \xmark                \\ \hline
Auto Balanced \cite{Dingautomatic}                     & \cmark                    & \cmark       & \cmark    & \xmark                \\ \hline
ThiNet \cite{thinet}                                              & \cmark                    & \cmark       & \xmark    & \xmark                \\ \hline
SqueezeNet \cite{squeezenet}                                          & \xmark                    & \cmark       & \xmark    & \cmark                \\ \hline
MobileNet \cite{mobilenet}                                           & \xmark                    & \xmark       & \xmark    & \cmark                \\ \hline
Zhang \cite{zhang2015efficient}                                               & \cmark                    & \cmark       & \xmark    & \xmark                \\ \hline
BinarizedNN \cite{binary}                                         & \xmark                    & \xmark       & \xmark    & \cmark                \\ \hline
XNORNet \cite{rastegari2016xnor}                                             & \xmark                    & \xmark       & \xmark    & \cmark                \\ \hline
Iannou \cite{ioannou2015training}                                              & \xmark                    & \xmark       & \cmark    & \cmark                \\ \hline
SNIP \cite{snip}                                                & \xmark                    & \xmark       & \cmark    & \xmark                \\ \hline
Lottery Ticket \cite{lottery_ticket}                                                & \cmark                    & \xmark       & \xmark    & \cmark                \\ \hline
Imp-Estimation \cite{importance_estimation}                                                & \cmark                    & \xmark       & \xmark    & \xmark                \\ \hline
NetTrim \cite{net_trim}                                                & \xmark                    & \xmark       & \cmark    & \cmark                \\ \hline
Runtime\_pruning \cite{runtime_pruning}                                                & \xmark                    & \cmark       & \cmark    & \xmark                \\ \hline
Learning-comp \cite{learning-compression}                                                & \cmark                    & \xmark       & \cmark    & \cmark                \\ \hline
Learned\_L0 \cite{learning_sparse}                                                & \xmark                    & \xmark       & \cmark    & \cmark                \\ \hline
Nucleus Nets \cite{nucleus}                                                & \xmark                    & \xmark       & \cmark    & \cmark                \\ \hline
Structure-learning \cite{structure_learning}                                                & \xmark                    & \cmark       & \cmark    & \xmark                \\ \hline
Discrimin-aware \cite{discrimination}                                                & \cmark                    & \cmark       & \cmark    & \xmark                \\ \hline
Structure-binary \cite{structuredseg}                                                & \xmark                    & \cmark       & \cmark    & \cmark                \\ \hline
Our Method                                           & \xmark                    & \xmark       & \xmark    & \xmark                \\ \hline
\end{tabular}
\vspace{0.1cm}
\caption{Summary of comparison of various pruning techniques. Loop 1 refers to finding empirical thresholds for pruning within a layer. Loop 2 accounts for iterations over layers as shown in Fig. \ref{fig:std_cht}. The last two columns refer to some specialized training procedures and changes to the network architecture or a requirement of custom architecture respectively.}
\label{table_comparison}
\end{table}

\section{Previous Work on Model Compression}
We divide model compression methods into four broad categories. The first are techniques that \textbf{prune individual weights}, such as in \cite{han2015deep}, \cite{lecun1990optimal}, \cite{AxNN} and \cite{optimalbrainsurgeon}. These techniques result in unstructured sparsity that is difficult to leverage in hardware. It requires custom hardware design and limits the savings that can be achieved. The second category tackles this problem by \textbf{removing entire filters}. Authors of \cite{li2016pruning}, \cite{molchanov2016pruning} and \cite{Dingautomatic} focus on finding good metrics to determine the significance of a filter and other pruning criterion. Authors of \cite{thinet} pose pruning as a subset selection problem based on the next layer's statistic. Methods in this category that do not compromise accuracy significantly require iterative retraining, incurring a heavy computational and time penalty on model design. While authors of \cite{molchanov2016pruning} analyze all layers together, their layer-wise analysis requires many iterations. Authors of \cite{discrimination} also remove filters by introducing multiple losses for each layer that select the most discriminative filters, but their method iterates until a stopping condition is reached within a layer and iterates over each layer, thus keeping both loops active. The third category, and the one that relates to our method the most, involves techniques that find different ways to \textbf{approximate weight matrices}, either with lower ranked ones or by quantizing such as in  \cite{denton2014exploiting}, \cite{jaderberg2014speeding}, \cite{squeezenet}, \cite{mobilenet} and \cite{zhang2015efficient}. However, these methods are done iteratively for each layer, preserving at least one of the loops in Fig. \ref{fig:std_cht}, making it a bottleneck for large network design. Authors of \cite{zhang2015efficient} use a similar idea, but choose a layer-wise rank empirically, by minimizing the reconstruction error. Authors of \cite{structuredseg} group the network into binarized segments and train these segments sequentially, resulting in the second loop, though over segments rather than layers. The binary bases of the segments are found empirically, and the method requires custom hardware and training methods. Authors of \cite{comp-aware} also use a similar scheme, but most of their savings appears to come from the regularizer rather than the post-processing. The regularization procedure adds more hyper-parameters to the training procedure, thus increasing iterations to optimize model design. Another difference is that we optimize both width and depth, and then train a new network from scratch, letting the network recreate the requisite filters. Aggressive quantization, all the way down to binary weights such as in \cite{binary} and \cite{rastegari2016xnor} results in accuracy degradation for large networks. The fourth category is one that learns the \textbf{sparsity pattern during training}. This includes modifying the loss function to aid sparsity such as in \cite{learning_sparse} and \cite{learning-compression}, or interspersing sparsifying iterations with the training iterations such as in \cite{learning-compression}. These result in extensive sparsity, but require longer training or non-standard training algorithms such as in \cite{net_trim}, \cite{runtime_pruning}, \cite{structure_learning} and \cite{structure_learning}, and do not always guarantee structured sparsity that can be quickly leveraged in existing hardware.  A non-standard architecture is created in \cite{nucleus} which consists of a well connected nucleus at initialization and the connectivity is optimized during training. Almost all of these works target static architectures, and optimize connectivity patterns rather than coming up with a new efficient architecture that does not require specialized training procedures or hardware implementations. Many of them are iterative over layers or iterative within a layer \cite{lottery_ticket}, \cite{importance_estimation}. None of these works optimize the number of layers.

\par There are \textbf{other techniques} that prune a little differently. Authors of \cite{nisp} define an importance score at the last but one layer and backpropagate it to all the neurons. Given a fixed pruning ratio, they remove the weights with scores below that. While this does get a holistic view of all layers, thus removing the second loop in Fig. \ref{fig:std_cht}, the first loop however still remains active as the pruning ratio is found empirically. Authors of \cite{ioannou2015training} learn a set of basis filters by introducing a new initialization technique. However, they decide the structure before training, and the method finds a compressed representation for that structure. The savings come from linear separability of filters into 1x3 and 3x1 filter, but the authors do not analyze if all the original filters need to be decomposed into separable filters. In comparison, our work takes in a trained network, and outputs an optimized architecture with reduced redundancy, given an accuracy target from the parent network. The work in \cite{snip} shows interesting results for pruning that do not require pre-training, but they too assume a fixed level of sparsity per layer.  The algorithm in \cite{hinton2015distilling} works a little differently than the other model compression techniques. It creates a new smaller student model and trains it on the outputs of the larger teacher model. However, there is no guarantee that the smaller model itself is free of redundancy, and the work does not suggest a method for designing the smaller network. Along with these differences, to the best of our knowledge, none of the prior works demonstrate a heuristic to optimize depth of a network. Another element of difference is that many of these methodologies are applied to weights, but we analyze activations, treating them as samples of the responses of weights acting on inputs. This is explained in detail in section III. 

\par The method we propose differs from these techniques in three major ways: a) the entire network with all layers is analyzed in a single shot, without iterations. Almost all prior works preserve at least one of the two loops shown in Fig. \ref{fig:std_cht}, while our methodology does not need either, making it a computationally viable way to optimize architectures of trained networks, b) the underlying choice of threshold is explainable and therefore exposes a knob to trade off accuracy and efficiency gracefully that can be utilized in energy constrained design tasks, and c) it targets both depth of the network and the width for all the layers simultaneously. To elaborate on the number of retraining iterations needed, let L represent the number of layers and N represent the number of iterations to find a suitable pruning threshold within a layer. Networks like VGG-16 have 13 convolutional layers, and for the sake of comparison, we assume a low estimate of N as 4, which means that 4 sparsity percentages are tested. Methods such as \cite{han2015deep}, \cite{denton2014exploiting} and \cite{molchanov2016pruning} that have both loop1 and loop2 active will require T*N=52 retraining iterations. Methods like \cite{lecun1990optimal}, \cite{importance_estimation} and \cite{learning-compression} that have only loop1 require N=4 number of retraining iterations and those that have only loop2 will require L=13 number of iterations, such as in \cite{comp-aware}, \cite{runtime_pruning} and \cite{structure_learning}. Since a whole retraining iteration can take many simulation days to converge, a large number of simulations is impractical. In contrast, our method only has 1 retraining iteration.

\par We would also like to point out that most of the works discussed in this section are orthogonal to ours and could potentially be used in conjunction, after using our method as a first order, lowest effort compression technique. These differences  are highlighted in tabular format in Table \ref{table_comparison}.

\begin{figure}[!t]
\begin{subfigure}{0.49\linewidth}
\centering 
\includegraphics[scale=.56]{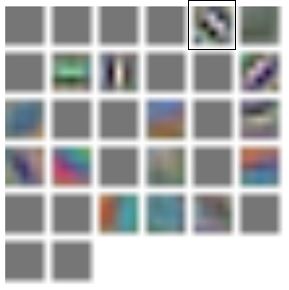} 
\caption{}
\label{fig:match1}
\end{subfigure}
\begin{subfigure}{0.49\linewidth}
\centering
\includegraphics[scale=.56]{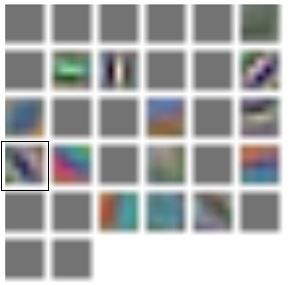}
\caption{}
\label{fig:match2}
\end{subfigure}
\caption{Visualization of pruning. The filter to be pruned is highlighted on the left and the filter it merges with is highlighted on the right. It can be seen that the merged filter incorporates the diagonal checkerboard pattern from the removed filter.}
\label{fig:prun_vis}
\end{figure}

\section{Framing Optimal Model Design as a Dimensionality Reduction Problem}
In this section, we present our motivation to draw away from the idea of ascribing significance to individual elements towards analyzing the space described by those elements together. We then describe how to use PCA in the context of CNNs and how to flatten the activation matrix to detect redundancy by looking at the principal components of this space. We then outline the method to analyze the results of PCA and use it to optimize the network layer-wise width and depth. The complete algorithm is summarized as a pseudo-code in Algorithm 1.
\par While this is not the first work that uses PCA to analyze models \cite{denton2014exploiting}, \cite{comp-aware}, the focus in this work is on a practical method of compressing pre-trained networks that does not involve multiple iterations of retraining. In other contexts, PCA has also been used to initialize neural networks \cite{pca-init}, and to analyze their adversarial robustness \cite{priya-exl}.

\subsection{Looking at the Filter Space Rather than Individual Filters}

\begin{figure}[!t]
\begin{subfigure}{0.475\textwidth}
\centering
\includegraphics[scale=.5]{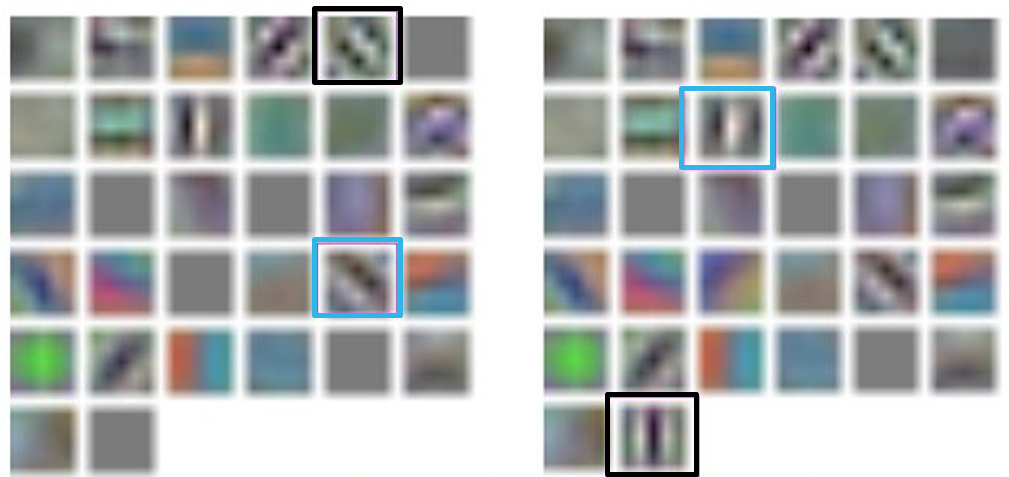} 
\caption{}
\label{fig:matches}
\end{subfigure}
\begin{subfigure}{0.475\textwidth}
\centering
\includegraphics[scale=.495]{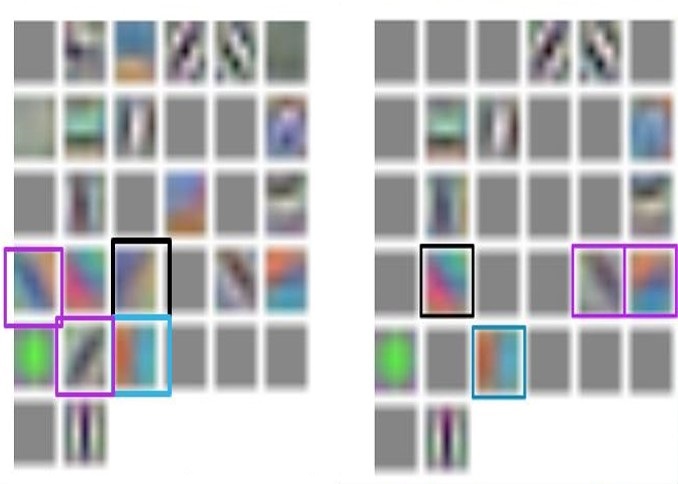}
\caption{}
\label{fig:mismatches}
\end{subfigure}
\caption{Fig. \ref{fig:matches}: Filter to be pruned is shown in black and the one that got changed the most is in blue. The filter in blue also had the highest Pearson correlation coefficient \cite{pearson} with the filter in black. Fig. \ref{fig:mismatches}: Mismatches are shown here. The filter that is pruned out is in black, the one closest to it according to Pearson coefficient is in blue. The two filters that changed the most after retraining are in pink.}
\label{fig:match_and_mis}
\end{figure}
\vspace{0.05in}

In an attempt to understand what happens during pruning and retraining, an exhaustive search was carried out during every iteration for a layer in order to identify the filter that caused the least degradation in accuracy upon removal. This means that at any iteration, all remaining filters were removed one at a time, and the net was retrained. Removal of whichever filter caused the least drop in accuracy was identified as the least significant filter for that iteration. This exhaustive analysis can only be carried out for small networks that do not require a long time to train. A small 3 layer network was trained on CIFAR-10 and visualized the effect of pruning and retraining. An animation was created from the iterative results of removing the identified least significant filter and retraining the model for the first layer, comprising of 32 filters. This analysis was carried out for the first layer so the filters can be effectively visualized. The resulting animation can be seen at this \href{https://github.com/isha-garg/Measure-Twice-Cut-Once/blob/master/exhaustive_reverse.gif}{link} \cite{pruning_gif_url}\, and gives a good insight into what occurs during pruning and retraining. Stills from the animation are shown in Fig. \ref{fig:prun_vis} and Fig. \ref{fig:match_and_mis}.

\par An interesting pattern is observed to be repeated throughout the animation: one or more of the remaining filters appear to `absorb' the characteristics of the filter that is pruned out. A particular snapshot is shown and explained in Fig. \ref{fig:prun_vis}. The filters before pruning are shown on the left in Fig. \ref{fig:match1}. The filter that is being removed in this iteration is highlighted. The filters after pruning and retraining are shown on the right in Fig. \ref{fig:match2}. It can be seen that the checkerboard pattern of the filter that was pruned out gets pronounced in the filter highlighted in Fig. \ref{fig:match2} upon retraining. Before retraining, this filter looked similar to the filter being pruned, but the similarity gets more pronounced upon retraining. This pattern is repeated often in the animation, and leads to the hypothesis that as a filter is removed, it seems to be recreated in some other filter(s) that visually appear to be correlated to it. Since the accuracy did not degrade,  we infer that if the network layer consists of correlated filters, the network can recreate the significant filters with any of them upon retraining. 

\par Given that it visually appears that each pruned out filter is absorbed by the one `closest' to it, we tried to determine if we could successfully predict the retained filter that would absorb the pruned out filter. Pearson correlation coefficient was used to quantify similarity between filters. A filter was chosen to be pruned and it was checked if the filter that changed the most upon retraining the system was the one which had the maximum Pearson correlation coefficient with the filter being pruned out. The L2 distance between the filter before and after retraining was used as a measure of change. Fig. \ref{fig:matches} shows an example iteration in which the filter identified as closest to the pruned out filter, and the filter that changed the most upon retraining were the same. But more significantly, it was observed that there were a lot of cases where the identified and predicted filters did not match, as sometimes one filter was absorbed by multiple filters combining to give the same feature as the pruned out filter, although each of them had low correlation coefficients individually. An illustrating example of such a mismatch is explained in Fig. \ref{fig:mismatches}. 

\par Viewing compression from the angle that each network has learned some significant and non significant filters or weights implicitly assumes that there is a static significance metric that can be ascribed to an element. This is counter-intuitive as the element can be recreated upon retraining. Even thinking of pruning as a subset selection problem does not account for the fact that on retraining, the network can adjust its filters and therefore the subset from which selection occurs is not static. This motivates a shift of perspective on model compression from removal of insignificant elements (neurons, connections, filters) to analyzing the space described by those elements. From our experiments, it would appear to be more beneficial instead to look at the behavior of the space described by the filters as a whole and find its dimensionality, which is discussed in the subsequent subsections.

 \begin{figure}[!t]
\centering
\includegraphics[width=.9\linewidth]{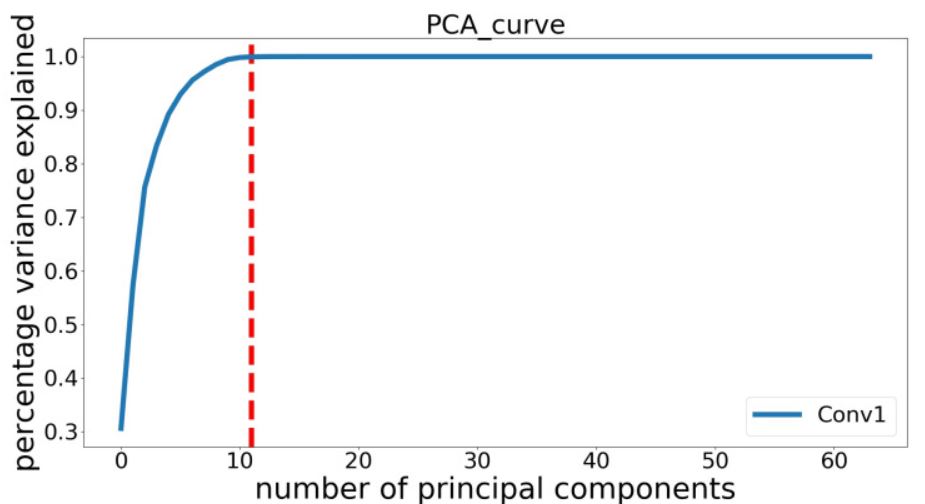}
%\vspace{0.1cm}
\caption{Cumulative percentage of the variance of input explained by ranked principal components. The red line identifies the significant dimensions that explain 99.9\% variance. }
\label{fig:pca_eg}
%\end{minipage}%
\end{figure}
\vspace{.05in}

\subsection{Analyzing the Space Described by Filters Using PCA}
\textbf{Principal Component Analysis (PCA):} Our method builds upon PCA, which is a dimensionality reduction technique that can be used to remove redundancy between correlated features in a dataset. It identifies new, orthogonal features which are linear combinations of all the input features. These new features are ranked based on the amount of variance of the input data they can explain. As an analogy, consider a regression problem to predict house rates with N samples of houses and M features in each sample. The input to PCA would be an NxM sized matrix, with N samples of M features among which we are trying to identify redundancy. 
\par \textbf{PCA for Dimensionality Reduction:} A sample output of PCA is shown in Fig. \ref{fig:pca_eg}, with cumulative explained variance sketched as a function of the number of top ranked principal components. The way this graph is utilized in the proposed method to uncover redundancy is by drawing out the red line, which shows how many features are required for explaining 99.9\% of the total variance in the input data. In this example, almost all the variance is explained by only 11 out of 64 features. Eleven new features can be constructed as linear combinations of the original 64 filters that suffice to explain virtually all the variance in the input, thus exposing redundancy in the feature space.

\begin{figure}[!t]
%\begin{minipage}[b]{.38\textwidth}
\centering
\includegraphics[width=.55\linewidth]{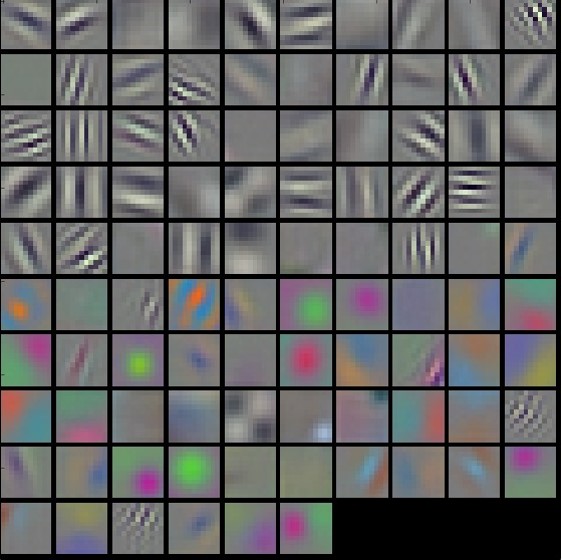}
\vspace{0.05cm}
\caption{First layer filters of AlexNet, trained on ImageNet. A considerable amount of redundancy is clearly visible in the filters.}
\label{fig:alexfil}
%\end{minipage}%
%\hspace{0.2cm}
%\begin{minipage}[b]{.61\textwidth}
\end{figure}

\par \textbf{PCA in the Context of CNNs:}
The success of currently adopted pruning techniques can be attributed to the redundancy present in the network. Fig. \ref{fig:alexfil} shows the filters of the first layer of AlexNet \cite{alexnet}. Many filters within the layer are highly correlated and potentially detect the same feature, therefore making insignificant contributions to accuracy. In the previous section, we deduced that pruned out filters, if redundant, could be recreated by a linear combination of retained filters without the retrained network suffering a drop in accuracy. This led us to view the optimal architecture as an intrinsic property of the space defined by the entire set of features, rather than of the features themselves. In order to remove redundancy,  optimal model design is framed as a dimensionality reduction problem with the aim of identification of the number of uncorrelated `eigenfilters' of the desired, smaller subspace of the entire hypothesis space of filters in a particular layer. By using PCA, the notion of the significance of a filter is implicitly removed, since the filters that are the output of PCA are linear combinations of all the original filters. It is the dimensionality which is of primary importance rather than the `eigenfilters'. We believe that the dimensionality determines an optimal space of relevant transformations, and the network can learn the requisite filters within that space upon training.

\begin{figure}[!t]
\centering
\includegraphics[scale=0.45]{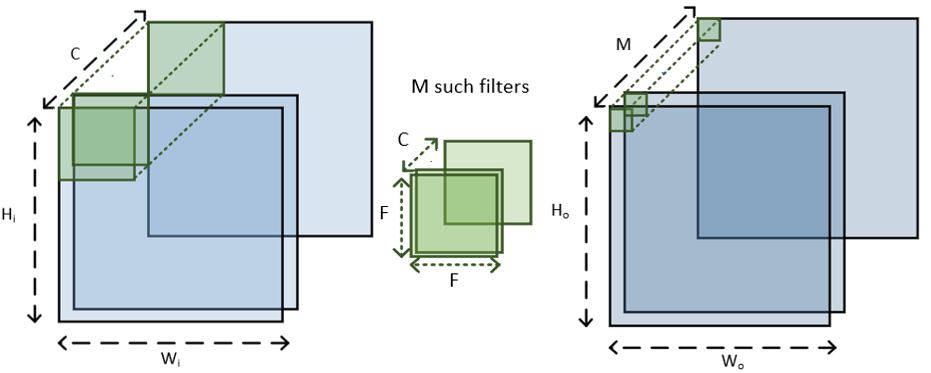}
\caption{The output of convolving one filter with an input patch can be viewed as the feature value of that filter. The green pixels in the output activation map make up one sample for PCA.}
\label{fig:featval}
%\end{minipage}%
\end{figure}

\par \textbf{Activations as Input Data to PCA for Detecting Filter Redundancy:}
The activations, which are instances of filter activity, are used as feature values of a filter to detect redundancy between the filters generating these activations. The standard input to PCA is a 2-dimensional matrix where each row represents a new sample, and each column corresponds to a feature value of a particular filter for all those samples. In this formulation, the feature value of a filter is its output value upon convolution with an input patch, as shown in Fig. \ref{fig:featval}. Hence a data point in the PCA matrix at the location [i,j] corresponds to the activation generated when the $i^{th}$ input patch is acted upon by the $j^{th}$ filter. The same input patch is convolved upon by all the filters, making up a full row of feature values for that input patch. As many of these input patches are available as there are pixels in one output activation map of that layer. Flattening the activation map after convolution gives many samples for all M filters in that layer. If there are activations that are correlated in this flattened matrix across all samples, it implies that they are generated by redundant filters that are looking at similar features in the input.

\begin{figure}[!t] 
\begin{subfigure}{0.485\linewidth}
\vspace{0.12cm}
\includegraphics[width=1\textwidth]{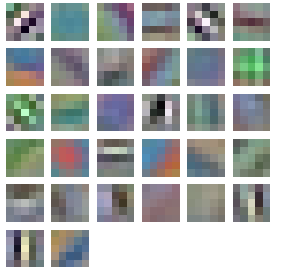} 
\caption{}
\label{fig:pre_trans}
\end{subfigure}
\begin{subfigure}{0.505\linewidth}
\includegraphics[width=1\textwidth]{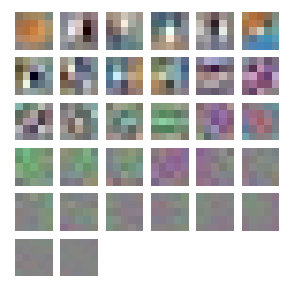}
\caption{}
\label{fig:post_trans}
\end{subfigure}
\vspace{0.2cm}
\caption{ The learned filters of a convolutional layer with 32 filters on CIFAR-10 are shown on the left and the corresponding ranked filters transformed according to principal components are shown on the right.}
\label{fig:pca_trans_filters}
\end{figure}

\par Let $A_{L}$ be the activation matrix obtained as the  output of a forward pass. \textit{L} refers to the layer that generated this activation map and is being analyzed for redundancy among its filters. The first filter-sized input patch convolves with the first filter to give the top left pixel of the output activation map. The same patch convolves with all M filters to give rise to a vector $\in \mathbb{R}^{1\times1\times M}$. This is viewed as one sample of M parameters, with each parameter corresponding to the activity of a filter. Sliding to the next input patch provides another such sample of activity. 
\par Suppose  $A_{L} \in \mathbb{R}^{N\times H\times W\times M}$, where N is the mini-batch size, H and W are the height and width of the activation map, and M is the number of filters that generated this map. Thus, it is possible to collect $N\times H\times W$ samples in one forward pass, each consisting of M parameters simply by flattening the matrix $A_{L} \in \mathbb{R}^{N\times H\times W\times M} \rightarrow B_{L} \in \mathbb{R}^{D\times M}$, where $D=N\times H\times W$. Since PCA is a data-intensive technique, we found that collecting data over enough mini batches such that $\frac{D}{M}$ is is roughly larger than 100 provides enough samples to detect redundancy. We then perform PCA analysis on $B_L$. We perform Singular Value Decomposition (SVD) on the  mean normalized, symmetric matrix $B_{L}^TB_{L}$ and analyze its M eigenvectors $\vec{v_{i}}$ and eigenvalues $\lambda_{i}$.  
\par The trace, tr($B_{L}^TB_{L}$) is the sum of the diagonal elements of the sample variance-covariance matrix, and hence equal to the sum of variance of individual parameters, which we call the total variance T.
$$tr(B_{L}^TB_{L})=\sum_{i=1}^{M}\sigma_{ii}^2 = T$$
The trace is also equal to the sum of eigenvalues. 
$$tr(B_{L}^TB_{L})=\sum_{i=1}^{M}\lambda_{i}$$
Hence, each $\lambda_{i}$ can be thought of as explaining a $\lambda_{i}/T$ ratio of total variance. Since the $\lambda_{i}$'s are ordered by largest to smallest in magnitude, we can calculate how many eigenvalues are cumulatively required to explain 99.9\% of the total variance, which we refer to as the significant dimensions for that layer,  $S_{L}$.  
$$S_{L}= \hat{M}:\frac{\sum_{i=1}^{\hat{M}}\lambda_{i}}{\sum_{i=1}^{M}\lambda_{i}} = 0.999$$

%This gives us a good way to approximate the original matrix in lower dimensions, by approximating $A_{L}$ as $\hat{A_{L}} \in  \mathbb{R}^{N*H*W*\hat{M}}$.
These significant dimensions are used to infer the optimized width and depth, as explained in the subsequent sections. From PCA we also know the transformation that was applied to $B_{L}$ and we can apply the same transformation to the filters to visualize the `principal' filters generated by PCA. This is shown in  Fig. \ref{fig:pca_trans_filters}. Fig. \ref{fig:pre_trans} shows the trained filters, and Fig. \ref{fig:post_trans} shows the ranked `eigenfilters' determined by PCA.  The filters are displayed according to diminishing variance contribution, with the maximum contributing component on top left and the least contributing component on the bottom right.

\begin{algorithm}[!t]
\caption{Optimize a pre-trained model}
\begin{algorithmic}[1]
    \Function{flatten}{num\_batches, layer}
        \For{batch = 1 to num\_batches}
            \State Perform a forward pass
            \State act\_layer  $\gets$ activations[layer]\Comment{size: N*H*W*C}
            \State reshape act\_layer into [N*H*W,C] 
            \For{sample in act\_layer}
                \State act\_flatten.append(sample) 
            \EndFor
        \EndFor
        \State \Return act\_flatten
    \EndFunction \\
%\end{algorithmic}

%\begin{algorithmic}
    \Function{run\_pca}{threshold, layer}
        %\Comment{run PCA and plot cumulative explained variance ratio and find number of significant dimensions for a layer}
        \State num\_batches $\gets \lceil(100*C/(H*W*N))\rceil$
        \State act\_flatten $\gets$ \Call{flatten}{num\_batches, layer}
        \State perform PCA on act\_flatten, C num\_components
        \State var\_cumul $\gets$ cumulative sum of explained\_var\_ratio
        \State pca\_graph $\gets$ plot var\_cumul against \#filters
        \State $S_L \gets$ \#components with var\_cumul$<$threshold
        \State \Return $S_L$
    \EndFunction \\
%\end{algorithmic}

%\begin{algorithmic}
    \Function{main}{threshold}
       \ForAll{layer in layers}
         \State $S_L \gets$ \Call{run\_PCA}{threshold, layer}
         \State S.append($S_L$)
       \EndFor
       \State new\_net $\gets$ [S[0]]
       \For{i $\gets$ 1 to num\_layers}
           \If{S[i]$>$S[$i-1$]}
               \State new\_net.append(S[i])
            \Else
                \State break
            \EndIf
       \EndFor
    \State new config: \# layers $\gets$ len(new\_net)
    \State each layer's \# filters $\gets S_L$ 
    \State randomly initialize a new network with new config
    \State train new network \Comment{Only training iteration}
    \EndFunction

 \end{algorithmic} 
 \end{algorithm}
 
\begin{figure*}[!t]
\centering
\begin{subfigure}{0.895\textwidth}
\centering
\includegraphics[width=1\linewidth]{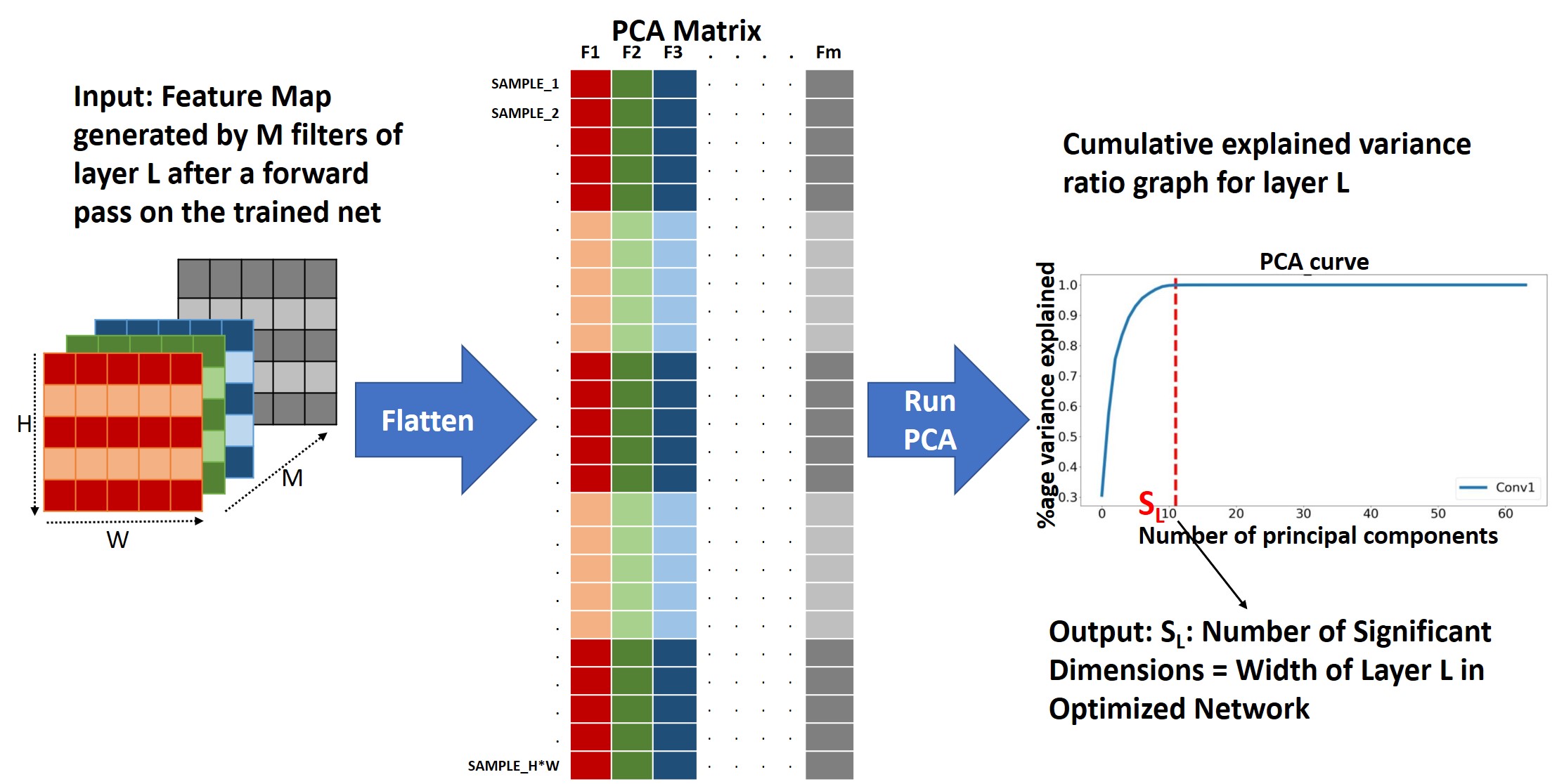} 
\caption{}
\label{fig:algoa}
\end{subfigure}
   
\begin{subfigure}{0.75\textwidth}
\centering
\includegraphics[width=1\linewidth]{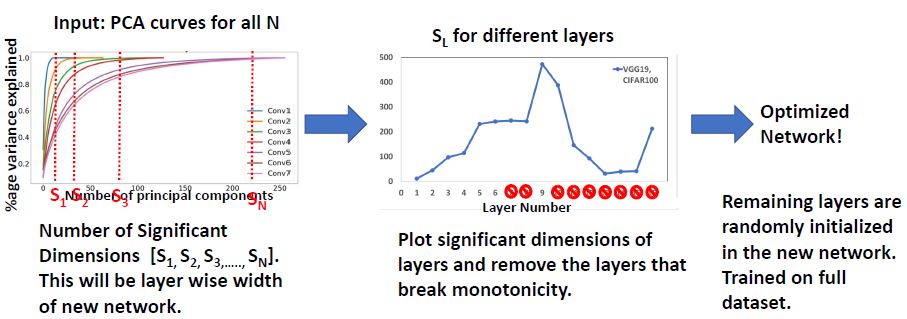}
\caption{}
\label{fig:algob}
\end{subfigure}
\caption{Visualization of the algorithm. Fig. \ref{fig:algoa} shows how to generate the PCA matrix for a particular layer, and then find its significant dimensions. Fig. \ref{fig:algob} shows how to use the results of Fig. \ref{fig:algoa} run in parallel on multiple layers to deduce the structure of the optimized network in terms of number of layers and number of filters in each layer. The structure of other layers (maxpool, normalization etc.) is retained from the parent network.}
\label{fig:algo}
\end{figure*}
\vspace{0.05in}

\subsection{Optimizing Width using PCA}
 The previous subsection outlines a way of generating PCA matrices for each layer. PCA analysis is then performed on these flattened matrices, and the cumulative variance explained is sketched as a function of the number of filters, as shown in Fig. \ref{fig:pca_eg}. The `significant dimensionality' of our desired space of filters is defined as the number of uncorrelated filters that can explain 99.9\% of the variance of features. This significant dimensionality, $S_L$ for each layer is the identified layer-wise width of the optimized network. Since this analysis can be performed simultaneously for all layers, one forward pass gives us the significant dimensions of all layers, which is used to optimize depth as explained in the next subsection.

\subsection{Optimizing Depth of the Network}
An empirical observation that emerged out of our experiments was a heuristic to optimize the number of layers of the neural network. A possible explanation for this heuristic could be arrived at by considering each layer as a transformation to progressively expand the input data into higher dimensions until the data is somewhat linearly separable and can be classified with desired accuracy. This means that the width of the network per layer should be a non-decreasing function of number of layers. However, as can be seen from the results in Section IV, summarized in Table \ref{table1}, the  number of significant dimensions expand up to a certain layer and then start contracting. We hypothesize that the layers that have lesser significant dimensions than the preceding layer are not contributing any relevant transformations of the input data, and can be considered redundant for the purpose of classification. If the significant dimensions are sketched for each layer, then the depth can be optimized by retaining the layers that maintain monotonicity of this graph. In Section IV, we show empirical evidence that supports our hypothesis by removing a layer and retraining the system iteratively. We notice that the accuracy starts degrading only at the optimized depth identified by our method, confirming that it is indeed a good heuristic for optimizing depth that circumvents the need for iterative retraining.

\par The methodology is summarized in the form of pseudo-code shown in Algorithm 1. The first procedure collects activations from many mini-batches and flattens it as described in the first part of Fig. \ref{fig:algoa}. It outputs a 2 dimensional matrix that is input to the PCA procedure in the second function. The number of components for the PCA procedure is equal to the number of filters generating that activation map. The second function, shown in the second part of Fig. \ref{fig:algoa} runs PCA on the flattened matrix and sketches the cumulative explained variance ratio as a function of number of components. It outputs the significant dimensions for that layer as the number of filters required to cumulatively explain 99.9\% of the total variance. In the third function, this process is repeated in parallel for all layers and a vector of significant dimensions is obtained. This is shown in Fig. \ref{fig:algob}. This corresponds to the width of each layer of the new initialized network. Next, we sketch the number of significant dimensions, and remove the layers that break the monotonicity of this graph. This decides the number of layers in the optimized network. The structure of the optimized network is hence obtained without any training iterations. The entire process just requires one training iteration (line 38 in Algorithm 1), on a new network initialized from scratch. This simplifies the pruning method considerably, resulting in a practical method to design efficient networks.

\subsection{Additional Insights}
\par Our method comes with two additional insights. First, the PCA graphs give designers a way to estimate the accuracy-efficiency tradeoff, and the ability to find a smaller architecture that retains less than 99.9\% of the variance depending on the constrained energy or memory budget of the application. Second, it offers an insight into the sensitivity of different layers to pruning.

\begin{figure}[!t]
\includegraphics[width=1\linewidth]{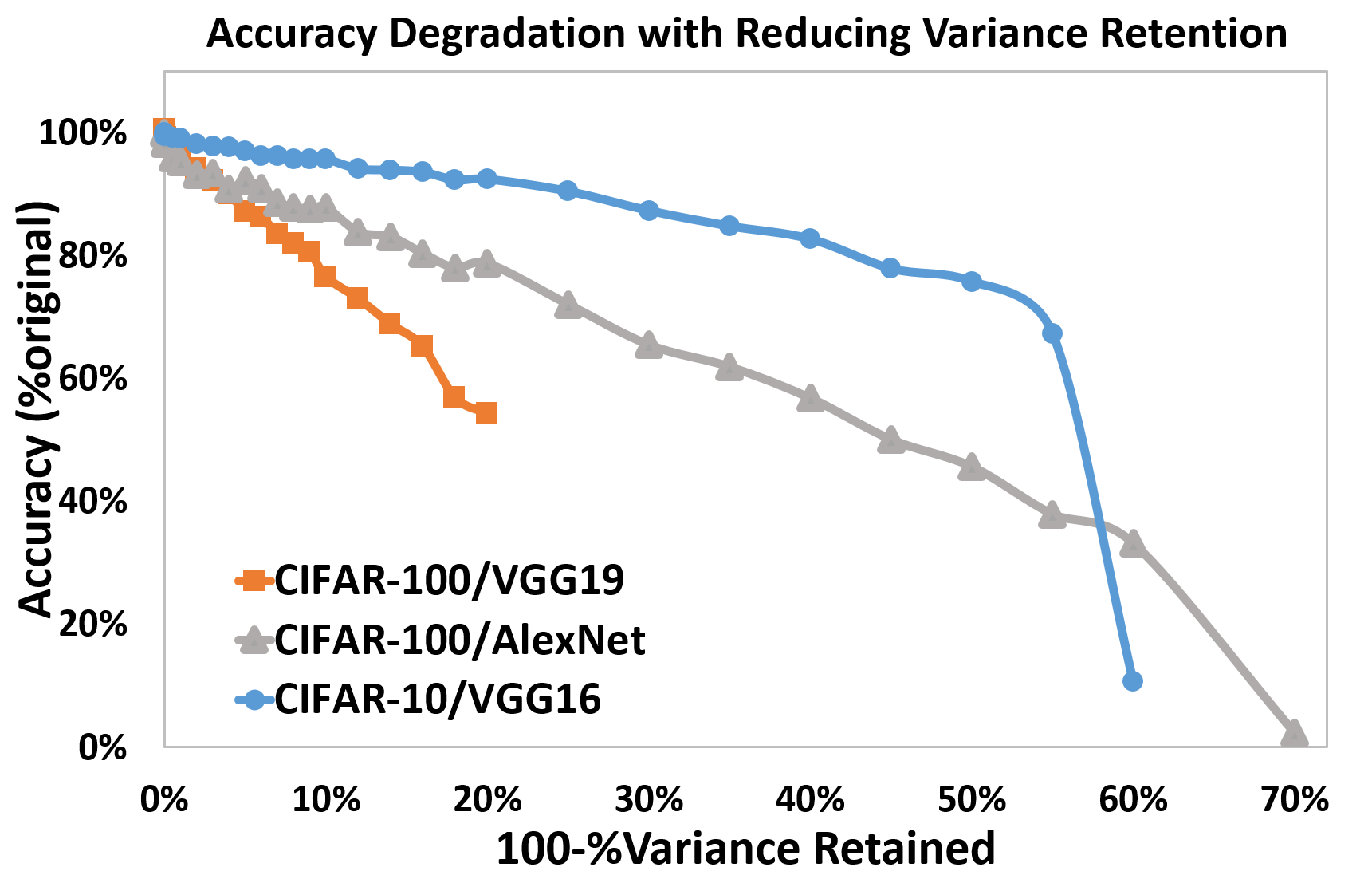} 
\caption{The degradation of accuracy w.r.t. target variance to explain for different networks. Each point here is a freshly trained network whose layer-wise width was decided by the corresponding amount of variance to explain on the x axis. The linearity of the graphs shows that reduction in variance retained is a good estimator of accuracy degradation.}
\label{fig:results}
\end{figure}
\par \textbf{Accuracy-Efficiency Tradeoff:} Fig. \ref{fig:results} shows the effect of decreasing the percentage of retained variance on accuracy for 3 different dataset-network pairs. Each point in the graph is a network that is trained from scratch, whose layer-wise architecture is defined by the choice of cumulative variance to retain, shown on the x-axis. The linearity of this graph shows that PCA gives a good, reliable way to arrive at an architecture for reduced accuracy without having to do empirical experiments each time. Section IV explains this figure in greater detail.
\par
\begin{figure}[!t]
\begin{subfigure}{1\linewidth}
\includegraphics[width=1\textwidth]{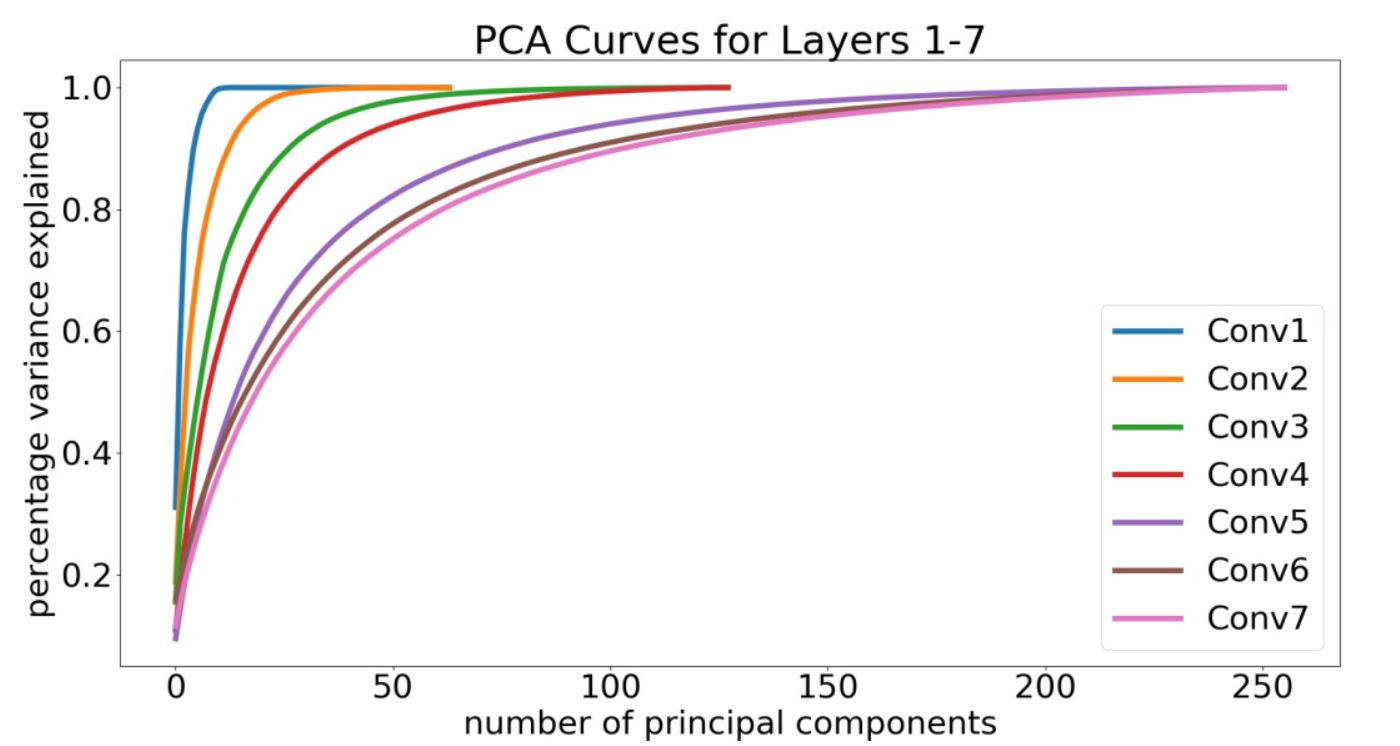} 
\caption{}
\label{fig:pca17}
\end{subfigure}
\begin{subfigure}{1\linewidth}
\includegraphics[width=1\textwidth]{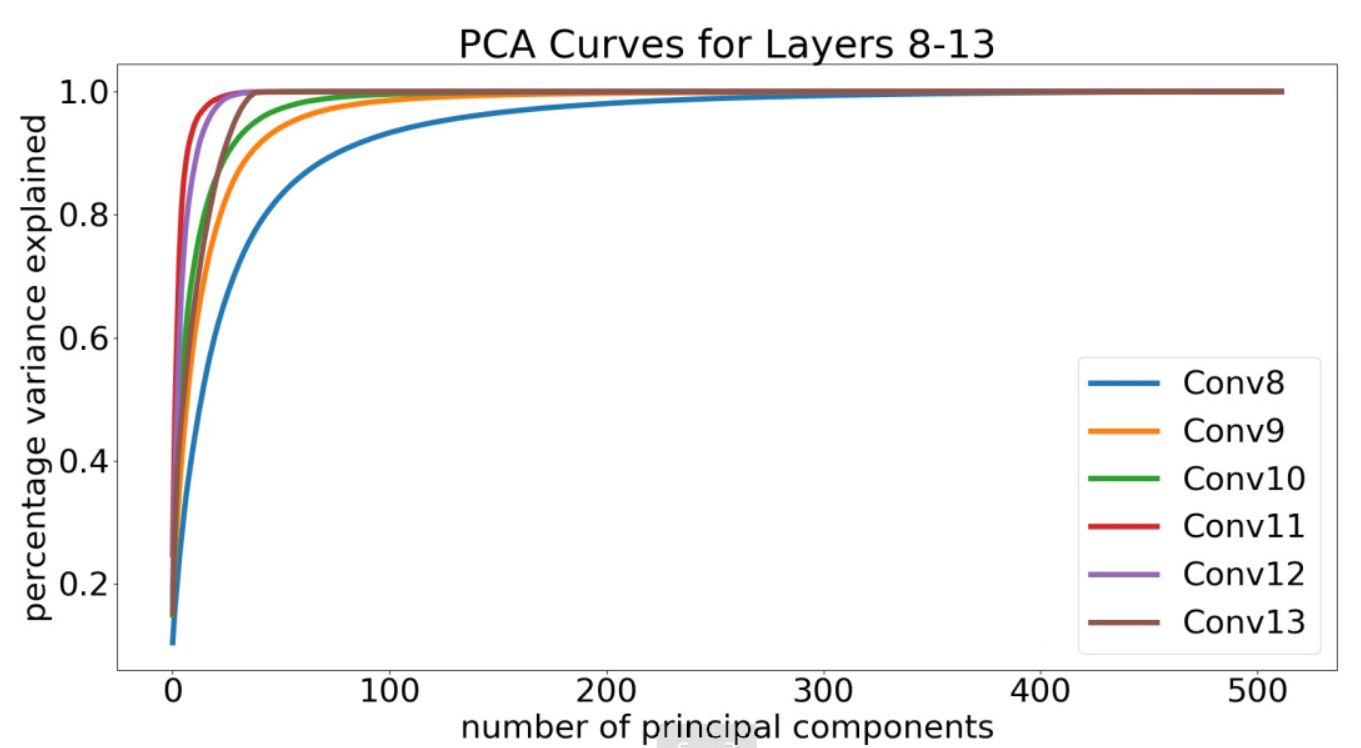}  
\caption{}
\label{fig:pca813}
\end{subfigure}
\caption{PCA graphs for different layers of CIFAR-10/VGG-16\_BN. Fig. \ref{fig:pca17} shows that layers 1-7 have increasing sensitivity to pruning, whereas the sensitivity decreases from layers 8 onwards as seen in Fig. \ref{fig:pca813}.}
\label{fig:sensitivity}
\end{figure}

 \textbf{Sensitivity of Different Layers to Pruning:} The second insight that the PCA graphs hint at is the sensitivity of layers to pruning, as a steeper graph points to the fact that lesser filters can explain most of the variance of the data. If the graph is very sharp, then it can be pruned more aggressively compared to a smoother graph where the principal components are well spread out with each component contributing to accuracy. This is shown in Fig. \ref{fig:sensitivity}, plotted for training VGG-16 \cite{VGG} adapted to CIFAR-10 data. From this figure, the expansion of the number of significant dimensions until layer 7 and subsequent contraction from layer 8 can also be observed, leading to the identification of the optimized number of layers before the classifier as 7.

\par Putting these two insights together helps designers with constrained energy budgets make informed choices of the architecture that gracefully trade off accuracy for energy. The final architecture depends only on the PCA graphs and the decided variance to retain. Therefore, given an energy budget, it is possible to identify a reduced amount of variance to retain across all layers that meets this budget. From the PCA graphs, the percentage of variance retained immediately identifies layer-wise width, and the depth can be identified from the contraction of the layer-wise widths. For even more aggressive pruning, the graphs expose the layers most resilient to pruning that can be targeted further. Note that all of these insights are available without a single retraining iteration. Thus a given energy budget can directly translate to an architecture, making efficient use of time and compute power.

\section{Results for Optimizing Network Structures}
Experiments carried out on some well known architectures are summarized in Table \ref{table1}. Discussions on the experiments, along with some practical guidelines for application are mentioned in the following subsections. PyTorch \cite{pytorch} was used to train the models, and the model definitions and training hyperparameters were picked up from models and training methodologies available at \cite{bearpaw}. A toolkit \cite{Dawood} available with PyTorch was used for profiling networks to estimate the number of operations and parameters for different networks.

\subsection{Experiments on Optimizing Width}
Fig. \ref{fig:sensitivity} shows the PCA graphs of different layers for all layers of the batch normalized VGG-16 network trained on CIFAR-10. These graphs are evaluated on the activations of a layer before the action of the non linearities, flattened as explained in Fig. \ref{fig:algo}. It can be observed that not all components are necessary to obtain 99.9\% cumulative variance of the input. The significant dimensions are identified, that is the dimensions needed to explain 99.9\% of the total variance, a new network is randomly initialized and trained from scratch with layer-wise width as identified by the significant dimensions. The resulting accuracy drops and savings in number of parameters and operations can be seen from the `Significant Dimensions' row of each network in Table \ref{table1}.

\subsection{Experiments on Optimizing Depth}
\par
\begin{figure}[!t]
        \begin{subfigure}{1\linewidth}
            \includegraphics[width=.95\textwidth]{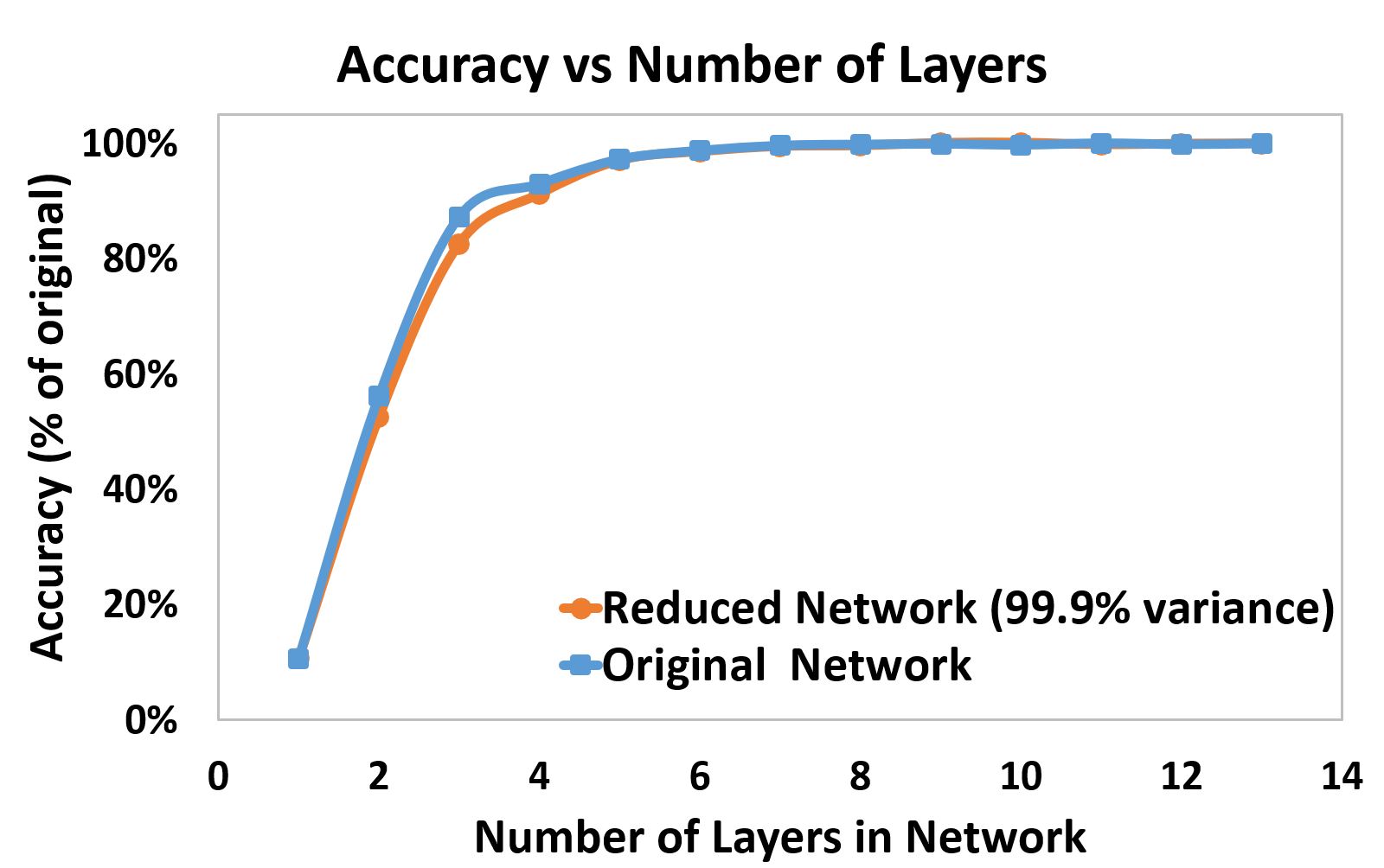}
            \caption{}%
            \label{fig:Acc_vs_layers}
        \end{subfigure}
        \begin{subfigure}{1\linewidth}   
            \includegraphics[width=.95\textwidth]{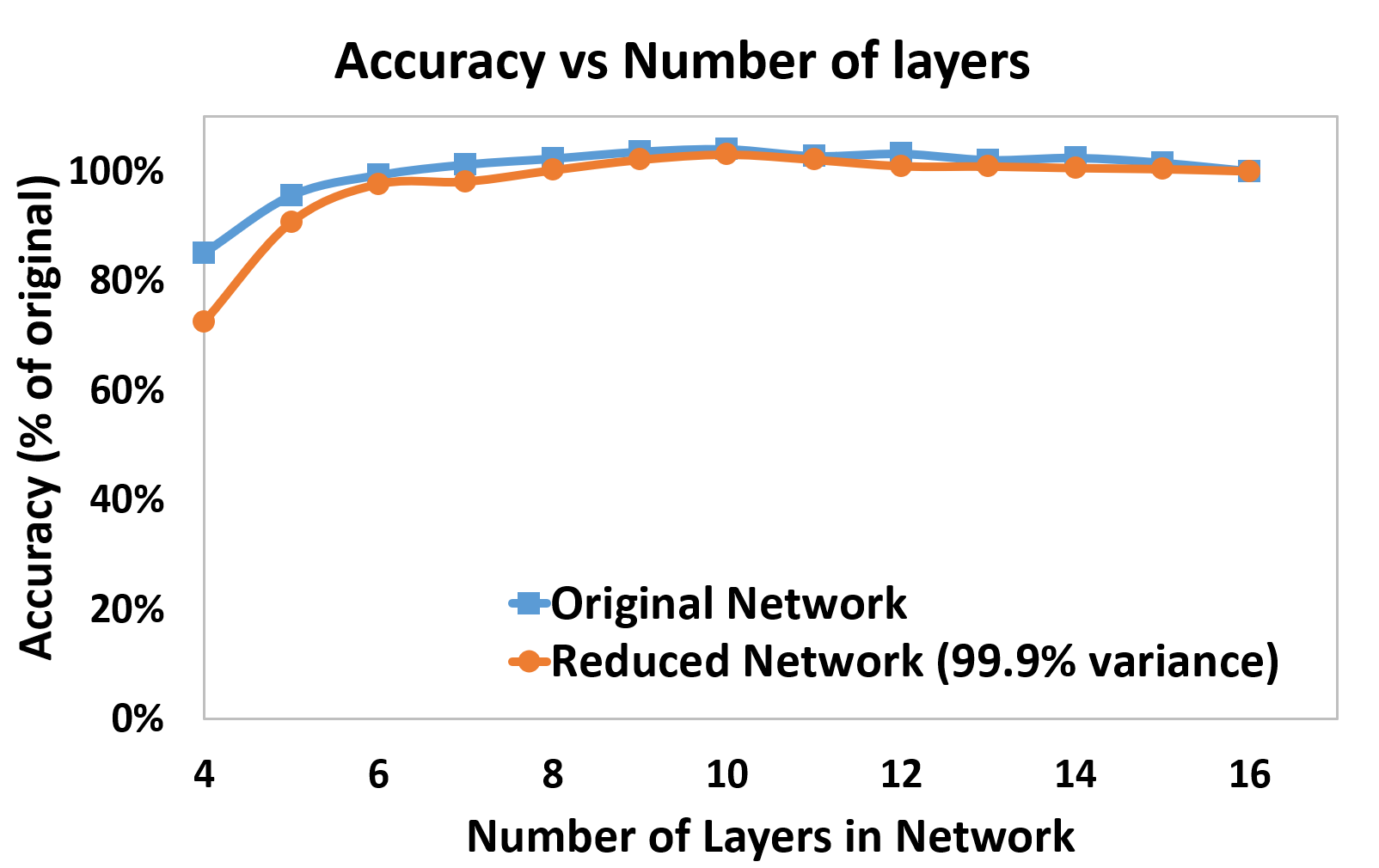}
            \caption{}%
            \label{fig:cifar100_layers}
        \end{subfigure}
        \vspace{0.2cm}
        \caption{The graphs illustrate how decreasing the number of layers affects accuracy. Fig. \ref{fig:Acc_vs_layers} shows results for CIFAR-10/VGG-16 and \ref{fig:cifar100_layers} for CIFAR-100/VGG-19. For CIFAR-100, going below 4 layers gave results highly dependent on initialization, so we only display results from layers 4 onwards.} 
        \label{fig:cifar}
\end{figure}

 Fig. \ref{fig:cifar} shows the degradation of retrained accuracy as the last remaining layer is iteratively removed, for CIFAR-10/VGG-16 and CIFAR-100/VGG-19. For both networks, a drop in accuracy is noticed upon removing the layers where the identified layer-wise width is still expanding. For example, significant dimensions for CIFAR-10/VGG-16 from Table \ref{table1} can be seen to expand until layer 7, which is the number of layers below which the accuracy starts degrading in Fig. \ref{fig:Acc_vs_layers}. A similar trend is observed for CIFAR-100/VGG-19, confirming that the expansion of dimensions is a good criterion for deciding how deep a network should be.

\subsection{Architectures with Reduced Accuracy}
\par The correlation between the explained variance and accuracy is illustrated in Fig. \ref{fig:results}. It shows results for CIFAR-10/VGG-16 and AlexNet and VGG-19 adapted to CIFAR-100. The graph shows how the accuracy degrades with retaining a number of filters that explain a decreasing percentage of variance. Each point refers to the accuracy of a new network trained from scratch. The configuration of the network was identified by the corresponding percentage of variance to explain, shown on the x-axis. The relationship is approximately linear until the unsaturated region of the PCA graphs is reached, where each filter contributes significantly to the accuracy. The termination point for these experiments was either when accuracy went down to random guessing or the variance retention identified a requirement of zero filters for some layer. For instance, the graph for CIFAR-100/VGG-19 stops at 80\% variance retention because going below this identified zero filters in some layers. The linearity of this graph shows that the explained variance is a good knob to tune for exploring the accuracy-energy tradeoff.

\subsection{Results and Discussion}
Putting together the ideas discussed, the results of employing this method on some standard networks are summarized in Table \ref{table1}. Each configuration is shown as a vector. Each component in the vector corresponds to a layer, with the value equal to the number of filters in a convolutional layer, and `M' refers to a maxpool layer. There are 5 network-dataset combinations considered, CIFAR-10/VGG-16, CIFAR-10/MobileNet, CIFAR-100/VGG-19, ImageNet/AlexNet and ImageNet/VGG-19. The row for significant dimensions just lists out the number of filters in that layer that explain 99.9\% of the variance. This will make up the layer-wise width of the optimized architecture. If these dimensions contract at a certain layer, then the final configuration has the contracting layers removed, thus optimizing the depth of the network. The table also shows the corresponding computational efficiency achieved, characterized by the reduction in number of parameters and operations. 

\par
\begin{figure}[!t]
        \begin{subfigure}{1\linewidth}
            \includegraphics[width=1\textwidth]{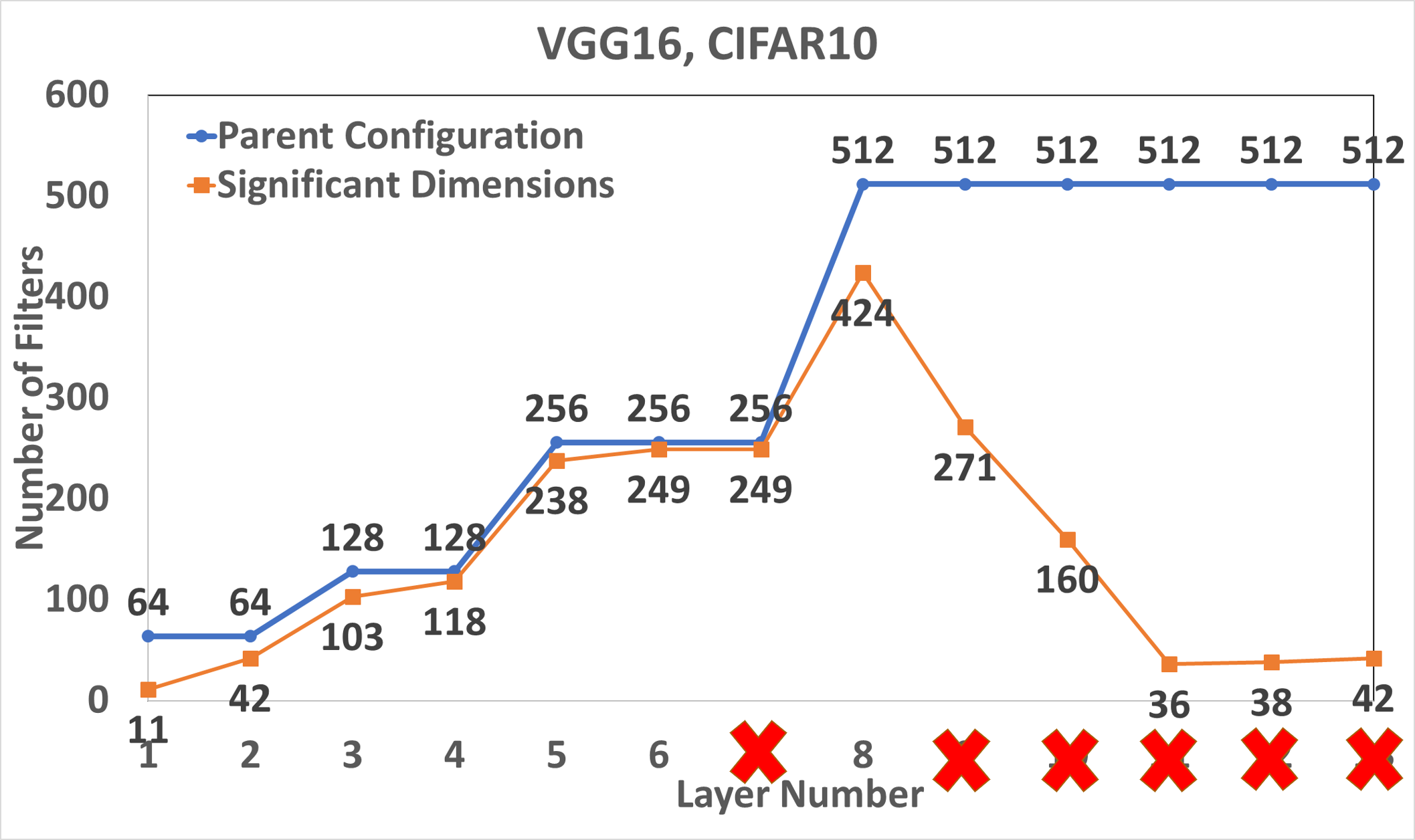}
            \caption{}%
            \label{fig:cifar10pic}
        \end{subfigure}
        \begin{subfigure}{1\linewidth}   
            \includegraphics[width=1\textwidth]{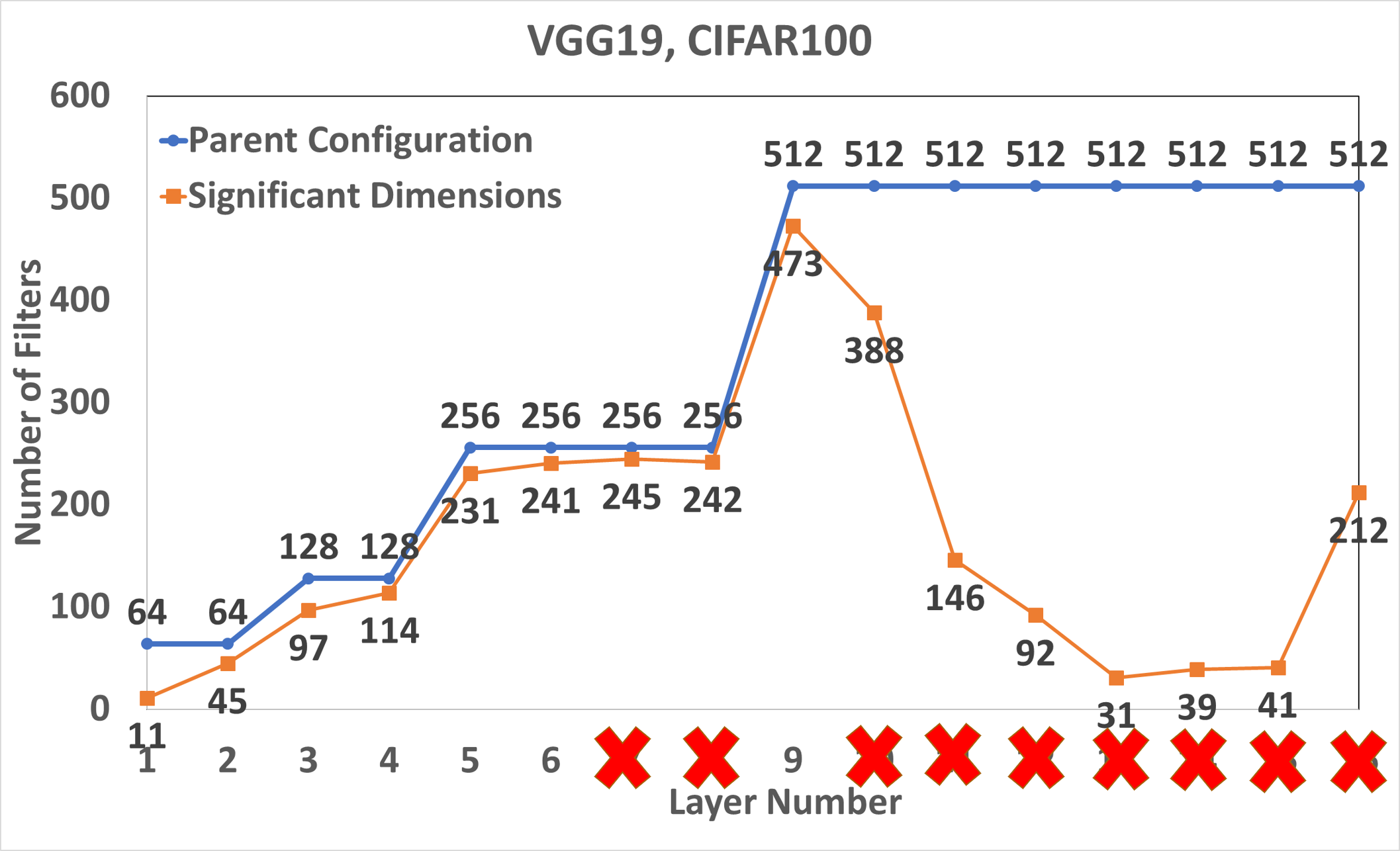}
            \caption{}%
            \label{fig:cifar100pic}
        \end{subfigure}
        \vspace{0.2cm}
        \caption{The graphs show how the significant dimensions vary with layers. The layers that do not result in a monotonic increase can be dropped, and are crossed out on the X axis. Fig. \ref{fig:Acc_vs_layers} shows results for CIFAR-10/VGG-16 and Fig. \ref{fig:cifar100_layers} for CIFAR-100/VGG-19.} 
        \label{fig:cifar2}
\end{figure}

 \textbf{CIFAR-10, VGG-16\_BN:}
The batch normalized version of VGG-16 was applied to CIFAR-10. The layer-wise significant dimensions are shown in Table \ref{table1}. From the table, it can be seen that in the third block of 3 layers of 256 filters, the conv layer right before maxpool does not add significant dimensions, while the layer after maxpool has an expanded number of significant dimensions. Hence, the conv layer before maxpool was removed while retaining the maxpool layer. In the third block, only one layer expands the significant dimensions, so it was retained and all subsequent layers were removed. The final network halved the depth to 7 layers with only a 0.7\% drop in accuracy. The resulting significant dimensions and layer removal is visualized in Fig. \ref{fig:cifar10pic}. The number of operations and parameters are reduced by 1.9X and 3.7X respectively.

\par \textbf{CIFAR-10, MobileNet:}
The MobileNet \cite{mobilenet} architecture was trained on CIFAR-10. The configuration is shown in Table \ref{table1}. The original block configuration is shown as a 3-tuple, with the first and second element corresponding to the number of filters in the first and second convolutional layers respectively, and the third element corresponds to the stride of the second convolutional layer. However, since the first convolutional layer of each block in MobileNet has a separate filter acting on each channel, we can not directly apply PCA to the intermediate activation map. As a workaround, a new network with the same configuration was trained, but with the first convolutional layer instantiated without grouping. This means that all the filters have the same depth as the input map and all channels of the input map are connected to all channels of the filters, as in standard convolution without any grouping. This results in increasing the number of operations and parameters by 8.5X and 8.7X respectively. The compression algorithm is instead applied to this network, as a proxy to the original network, and a reduction of 14.5X and 34.4X is seen in  the number of operations and parameters respectively, which translates to a reduction of 1.7X and 3.8X compared to the original MobileNet architecture (with grouping). The blocks are all reduced to just one convolutional layer, with no grouping. Even though the original network is considered to be one of the most efficient networks, we are able to further reduce its size and computational complexity while gaining almost a percentage point in accuracy. The number of layers reduce from 28 in the original network to 8 in the optimized network.

\definecolor{Gray}{gray}{0.9}
\begin{table*}[!t]
\scalebox{0.9}{\begin{tabular}{@{}|c|c|c|c|c|@{}}
\toprule
\multicolumn{1}{|c|}{} & \multicolumn{1}{c|}{CONFIGURATION} & \multicolumn{1}{c|}{ACCURACY} & \multicolumn{1}{c|}{\#OPS} & \multicolumn{1}{c|}{\#PARAMS} \\ \midrule
\rowcolor{Gray}
\textbf{} & \multicolumn{1}{c}{\textbf{Dataset, Network: CIFAR-10, VGG-16\_BN}} & \multicolumn{1}{c}{} & \multicolumn{1}{c}{} & \multicolumn{1}{c|}{}\\ \hline
\multicolumn{1}{|c|}{\textbf{Initial Config.}} & \multicolumn{1}{c|}{{[}64, 64, `M', 128, 128, `M', 256, 256, 256, `M', 512, 512, 512, `M', 512, 512, 512{]}} & \multicolumn{1}{c|}{94.07\%} & \multicolumn{1}{c|}{1X} & \multicolumn{1}{c|}{1X} \\ \midrule
\multicolumn{1}{|c|}{\textbf{Sig. Dimensions}} & \multicolumn{1}{c|}{{[}11, 42, `M', 103, 118, `M', 238, 249, 249, `M', 424, 271, 160, `M', 36, 38, 42{]}} & \multicolumn{1}{c|}{93.76\%} & \multicolumn{1}{c|}{1.9X} & \multicolumn{1}{c|}{3.7X} \\ \midrule
\multicolumn{1}{|c|}{\textbf{Final Config.}} & \multicolumn{1}{c|}{{[}11, 42, `M',  103, 118, `M', 238, 249, `M', 424, `M'{]}} & \multicolumn{1}{c|}{93.36\%} & \multicolumn{1}{c|}{2.9X} & \multicolumn{1}{c|}{7.7X} \\ \midrule

\rowcolor{Gray}
\textbf{} & \multicolumn{1}{c}{\textbf{Dataset, Network: CIFAR-100, VGG-19\_BN}} & \multicolumn{1}{c}{} & \multicolumn{1}{c}{} & \multicolumn{1}{c}{}\\ \midrule
\multicolumn{1}{|c|}{\textbf{Initial Config.}} & \multicolumn{1}{c|}{{[}64,  64, `M', 128, 128, `M', 256, 256, 256, 256, `M', 512, 512, 512, 512, `M',  512, 512, 512, 512, `M'{]}} & \multicolumn{1}{c|}{72.09\%} & \multicolumn{1}{c|}{1X} & \multicolumn{1}{c|}{1X} \\ \midrule
\multicolumn{1}{|c|}{\textbf{Sig. Dimensions}} & \multicolumn{1}{c|}{{[}11, 45, `M',  97, 114, `M', 231, 241, 245, 242, `M', 473, 388, 146, 92, `M', 31, 39, 42,  212, `M'{]}} & \multicolumn{1}{c|}{71.59\%} & \multicolumn{1}{c|}{1.9X} & \multicolumn{1}{c|}{3.7X} \\ \midrule
\multicolumn{1}{|c|}{\textbf{Final Config.}} & \multicolumn{1}{c|}{{[}11, 45, `M',  97, 114, `M', 231, 245, `M', 473, `M'{]}} & \multicolumn{1}{c|}{73.03\%} & \multicolumn{1}{c|}{3.8X} & \multicolumn{1}{c|}{9.1X} \\ \midrule

\rowcolor{Gray}
\textbf{} & \multicolumn{1}{c}{\textbf{Dataset, Network: CIFAR-10, MobileNet}} & \multicolumn{1}{c}{} & \multicolumn{1}{c}{} & \multicolumn{1}{c|}{}\\ \midrule

\begin{tabular}[|c|]{@{}l@{}} \textbf{Initial Config:}\\ With grouping\\ (W/o grouping)\end{tabular} & \begin{tabular}[|c|]{@{}c@{}}{[}32, (64,64,1), (128,128,2), (128,128,1), (256,256,2), (256,256,1), (512,512,2), (512,512,1), \\ (512,512,1), (512,512,1), (512,512,1), (512,512,1), (1024,1024,2), (1024,1024,1){]}\end{tabular} & \begin{tabular}[|c|]{@{}c@{}}90.25\%\\ (92.17\%)\end{tabular} & \begin{tabular}[|c|]{@{}c@{}}1X\\ (1X)\end{tabular}    & \begin{tabular}[|c|]{@{}c@{}}1X\\ (1X)\end{tabular}     \\ \midrule
\textbf{Sig. Dimensions}                                                                         & \begin{tabular}[|c|]{@{}c@{}}{[}10, (24,21,1), (46,40,2), (103,79,1), (104,85,2), (219,167,1), (199,109,2), \\ (235,99,1), (89,10,1), (10,2,1), (10,2,1), (10,2,1), (4,4,2), (24,16,1){]}\end{tabular}                           & 91.33\%                                                     & \begin{tabular}[|c|]{@{}c@{}}1X\\ (8.5X)\end{tabular}    & \begin{tabular}[c]{@{}c@{}} 3.1X\\ (28.1X)\end{tabular} \\ \midrule
\textbf{Final config.}                                                                            & {[}10, (24,1), (46,2), (103,1), (104,2), (219,2), (235,2){]}                                                                                                                                                                   & 91.08\%                                                     & \begin{tabular}[c]{@{}c@{}}1.7X\\ (14.4X)\end{tabular} & \begin{tabular}[c]{@{}c@{}}3.9X\\ (34.4X)\end{tabular} \\ \midrule

\rowcolor{Gray}
\textbf{} & \multicolumn{1}{c}{\textbf{Dataset, Network: CIFAR-100, AlexNet}} & \multicolumn{1}{c}{} & \multicolumn{1}{c}{} & \multicolumn{1}{c|}{}\\ \midrule
\multicolumn{1}{|c|}{\textbf{Initial Config.}} & \multicolumn{1}{c|}{{[}64, 192, 384, 256, 256{]}} & \multicolumn{1}{c|}{42.77\%} & \multicolumn{1}{c|}{1X} & \multicolumn{1}{c|}{1X} \\ \midrule
\multicolumn{1}{|c|}{\textbf{Sig. Dimensions}} & \multicolumn{1}{c|}{{[}44,119,304,251,230{]}} & \multicolumn{1}{c|}{41.74\%} & \multicolumn{1}{c|}{1.6X} & \multicolumn{1}{c|}{1.5X} \\ \midrule
\multicolumn{1}{|c|}{\textbf{Final Config.}} & \multicolumn{1}{c|}{{[}44,119,304,251{]}} & \multicolumn{1}{c|}{41.66\%} & \multicolumn{1}{c|}{2.1X} & \multicolumn{1}{c|}{2.1X} \\ \midrule

\rowcolor{Gray}
\textbf{} & \multicolumn{1}{c}{\textbf{Dataset, Network: ImageNet, VGG-19\_BN}} & \multicolumn{1}{c}{} & \multicolumn{1}{c}{} & \multicolumn{1}{c|}{}\\ \midrule
\multicolumn{1}{|c|}{\textbf{Initial Config.}} & \multicolumn{1}{c|}{{[}64,  64, `M', 128, 128, `M', 256, 256, 256, 256, `M', 512, 512, 512, 512, `M',  512, 512, 512, 512, `M'{]}} & \multicolumn{1}{c|}{74.24\%} & \multicolumn{1}{c|}{1X} & \multicolumn{1}{c|}{1X} \\ \midrule
\multicolumn{1}{|c|}{\textbf{Sig. Dimensions}} & \multicolumn{1}{c|}{{[}6, 30, `M',  49, 100, `M', 169, 189, 205, 210, `M', 400, 455, 480, 490, `M', 492, 492,  492, 492, `M'{]}} & \multicolumn{1}{c|}{74.00\%} & \multicolumn{1}{c|}{1.7X} & \multicolumn{1}{c|}{1.1X} \\ \bottomrule
\end{tabular}}
\vspace{0.1cm}
\caption{Summary of Results}
\label{table1}
\end{table*}

\par \textbf{CIFAR-100, VGG-19\_BN:} 
The analysis was expanded to CIFAR-100, using the batch normalized version of VGG-19. A similar trend was seen as in the previous case, and the optimized network consisted of 7 layers again. The resulting significant dimensions and layer removal is visualized in Fig. \ref{fig:cifar100pic}. An increase in accuracy of nearly one percent was observed, presumably owing to the fact that the network was too big for the dataset, thus having a higher chance of overfitting. The final reduction in number of operations and parameters is 3.8X and 9.1X, respectively.

\par \textbf{CIFAR-100, AlexNet:} 
To change the style of architecture to one with a smaller number of layers, the analysis was carried out for CIFAR-100 dataset on the AlexNet architecture and it was observed that the layer-wise depth decreased for all layers, but not by a large factor, as AlexNet does not seem to be as overparametrized a network as the VGG variants. However, the last two layers could be removed as they did not expand the significant dimensions, resulting in a reduction of 2.1X in both the number of operations and parameters.

\par \textbf{ImageNet, VGG-19\_BN:} 
The final test was on the ImageNet dataset, and the batch normalized VGG-19 network was used. In the previous experiments, the VGG network adapted to CIFAR datasets had only one fully connected layer, but here the VGG network has 3 fully connected layers, which take up the bulk of the total number of parameters (86\% of the total parameters are concentrated in the three fully connected layers.) Since the proposed compression technique only targets the convolutional layers, the total reduction in number of parameters is small. However the total number of operations still reduced by 1.7X the original number with just a 0.24\% drop in accuracy. Here, the depth did not reduce further as the number of significant dimensions remained non decreasing, and therefore reducing layers resulted in an accuracy hit. The final configuration is the same as the width-reduced configuration shown in Table \ref{table1}.

\par \textbf{Limitations:} One of the major limitations of this method is that it does not apply to ResNet style networks with shortcut connections. Removing dimensions at a certain layer that connects directly to a different layer can result in recreating the significant dimensions in the latter layer, thus making this analysis incompatible. Similarly the method does not directly apply to layers with grouping, since the filters do not act on common inputs channels. The workaround, as discussed in the case of MobileNet in section IV, is to train a new network with the same configuration but without grouping, and apply the method to that network. Another limitation is that this method only applies to a pre-trained network. We do not claim that it results in the most optimal compressed architecture; instead, this is the lowest effort compression that is available at negligible extra cost and can be used as a first order, coarse grained compression method. More fine grained methods can then be applied on the resulting structure. Another point to note is that since we view the compression method as identification of relevant subspace of filters, we do not apply it to the fully connected layers. However, if there were many fully connected layers, the resulting activations are already flattened and our method for reducing width can still be applied in a straightforward manner. 
%Just comparing it to the convolutional layers reduces parameters to 0.82X   

\subsection{Some Practical Considerations for the Experiments}
Three guidelines were followed throughout all experiments. First, while the percentage variance one would like to retain depends on the application and acceptable error tolerance,it was empirically found that preserving 99.9\% is a sweet spot with about half to one percentage point in accuracy degradation and a considerable gain in computational cost. Second, this analysis has only been done on activation outputs for convolutional layers before the application of non-linearities such as ReLU. Non-linearities introduce more dimensions, but those are not a function of the number of filters in a layer. And lastly, the number of samples to be taken into account for PCA are recommended to be around 2 orders of magnitudes more than the width of the layer (number of filters to detect redundancy in). Note that one image gives height times width number of samples, so a few mini-batches are usually enough to gather these many samples. It is easier in the first few layers as the activation map is large, but in the later layers, activations need to be collected over many mini-batches to make sure there are  enough samples to run PCA analysis on. However, this is a fraction of the time and compute cost of running even a single test iteration (forward pass over the whole dataset), and negligible compared to the cost of retraining. There is no hyper-parameter optimization followed in these experiments; the same values as for the original network are used.

\section{Conclusion}
A novel method to perform a single shot analysis of any given trained network to optimize network structure in terms of both the number of layers and the number of filters per layer is presented. The analysis is free of iterative retraining, which reduces the computational and time complexity of pruning a trained network by a large number of retraining iterations. It has explainable results and takes the guesswork out of choosing layer-wise thresholds for pruning. It exposes an accuracy-complexity knob that model designers can tweak to arrive at an optimized design for their application, and highlights the sensitivity of different layers to pruning. It is applied to popular networks and datasets. At negligible extra time and computational cost of analysis, an optimized structure is identified that achieves up to 3.8X reduction in number of operations and up to 9.1X reduction in number of parameters, with less than 1\% drop in accuracy upon training on the same task. We apply the algorithm to a highly efficient network, MobileNet and are able to achieve a reduction of 1.7X and 3.9X in the number of operations and parameters respectively, while improving accuracy by almost one percentage point.

%\bibliographystyle{ieeetr}
%\bibliographystyle{nips_2018}

\ifCLASSOPTIONcaptionsoff
  \newpage
\fi



%\bibliographystyle{IEEEtran}
%\begin{thebibliography}{1}
\bibliographystyle{IEEEtran}
\bibliography{references}
%\end{thebibliography}


% \begin{IEEEbiography}{Isha Garg}
% \end{IEEEbiography}

% \begin{IEEEbiographynophoto}{Priyadarshini Panda}

% % insert where needed to balance the two columns on the last page with
% %\newpage

% \begin{IEEEbiographynophoto}{Kaushik Roy}


%\vfill


\EOD

\end{document}